\documentclass[german,english]{bvm2020}

\def\articlenumber{2772}

\title{Prediction of MRI Hardware Failures based on Image Features using Time Series Classification}

\titlerunning{Time Series Classification}

\author{Nadine Kuhnert$^1$, Lea Pfl\"uger$^1$, Andreas Maier$^1$}

\authorrunning{Kuhnert, Pfl\"uger \& Maier}

\institute{$^1$Pattern Recognition Lab, Friedrich-Alexander University Erlangen-Nuremberg}

\email{nadine.kuhnert@fau.de}

\begin{document}

%
\selectlanguage{english}

\maketitle

\begin{abstract}
Already before systems malfunction one has to know if hardware components will fail in near future in order to counteract in time. Thus, unplanned downtime is ought to be avoided. In medical imaging, maximizing the system's uptime is crucial for patients' health and healthcare provider's daily business. We aim to predict failures of Head/Neck coils used in Magnetic Resonance Imaging (MRI) by training a statistical model on sequential data collected over time. As image features depend on the coil's condition, their deviations from the normal range already hint to future failure.
Thus, we used image features and their variation over time to predict coil damage. 
After comparison of different time series classification methods we found Long Short Term Memorys (LSTMs) to achieve the highest F-score of 86.43\% and to tell with 98.33\% accuracy if hardware should be replaced.
\end{abstract}

\section{Introduction}
Often it feels like hardware failures occur all of a sudden. However, data contain deviations from the normal range already before breakage and carry hints of future failing parts. In Magnetic Resonance Imaging (MRI), systems are extensively used and unplanned downtimes come with high costs. High image quality and seamless operation are crucial for the diagnostic value. Radiofrequency coils are essential hardware as they receive signals which form the basis of the desired diagnostic image~\cite{maier2018medical}. Thus, the goal is to prevent unplanned coil failure by exchanging or repairing the respective coil before its malfunction. 
In this work, we predict failures of Head/Neck coils using image-related measurements collected over time. Therefore, we aim to solve a time series classification (TSC) problem. 

In literature, time series classification is tackled by a range of traditional Machine Learning (ML) algorithms such as Hidden Markov Models, Neural Networks or Linear Dynamic Systems~\cite{bishop2006pattern}. 
Wang et al.~\cite{wang2017time} are the first who introduced Convolutional Neural Networks (CNN) for the classification of univariate time series where no local pooling layers are included and thus, the length of time series kept the same for all convolutions. They also applied a deep Residual Network (ResNet) for TSC. The main characteristic of a ResNet is the addition of a linear shortcut that connects the output of a residual block to its input. Furthermore, Recurrent Neural Networks (RNNs) contain loops to allow the network to store previous information which is essential for time series applications. A special kind of RNNs is the Long Short-Term Memory (LSTM) introduced by~\cite{hochreiter1997long} to specifically incorporate long-term dependencies. Deep neural networks are powerful but often struggle with overfitting and long computation times which one can counteract using the dropout technique~\cite{srivastava2014dropout}.

The task of hardware failure prediction using time series data has not been addressed widely in literature.
Lipton et al.~\cite{lipton2015learning} used LSTMs in order to predict diagnosis based on clinical, time series data. Furthermore, prediction of high performance computing system failures using sequential data was trained using Support Vector Machines~\cite{mohammed2019failure}. Jain et al.~\cite{jain2019image} found that hardware failures can be predicted based on image features, but did not examine collections of image features over time.

\section{Materials and methods}
In order to prevent malfunction, we applied different ML methods to determine broken coils. Firstly, we describe the data and available features. Secondly, we depict preprocessing steps, present the applied models and their configuration.
\subsection{Data}
We employ classification algorithms on data which was acquired by 238 Siemens MAGNETOM Aera 1.5T MRI systems. Data was collected before every examination using a 20-channel Head/Neck coil since May 1st 2019 all over the world, as well as, from measurements performed at Siemens' research halls specifically generating data sets for various hardware failure cases. This yields 29878 sequences in total which contain 2.2\% sequences of defective coils.
We derive image features from coil adjustment measurements which are generated before the clinical scan. Thus, reconstruction of any patient-specific features is impossible and guarantees non-clinical, fully anonymized data. 

We use image features of coil elements reported by MRI systems  which are represented by four continuous numerical measurements per time instance:
\begin{itemize}
\item[]\textit{Channel Signal Noise Level (CNL)} 
The coil noise is measured every time a new coil
configuration is selected or the patient table is moved. This happens at least once per
examination. After the noise measurement the noise level is calculated and reported as one value for each
channel.
\item[]\textit{Channel Signal to Noise Ratio (CSP)}
During coil sensitivity adjustments, both coil noise
and sensitivity are measured. Depending on the coil element the signal to noise ratio
is estimated.
\item[]\textit{Channel Signal to Signal (SSR)}
During the adjustment pre-scanning process also body
coil measures are performed. SSR uses the signal measure of the Body coil and the signal
around the isocenter of local Head/Neck coil to calculate a signal ratio between Body
coil and Head/Neck coil element.
\item[]\textit{Channel Signal to Noise Ratio at Isocenter (CSI)}
CSI combines the channel signal to
noise ratio (CSP) with the channel signal in the isocenter and reports the respective ratio.
\end{itemize}

\subsection{Preprocessing}
First, we centered and scaled the data of each individual feature independently by subtracting the mean and dividing by the standard deviation, respectively. 
Furthermore, we  artificially produced new defective time series based on training data set in order to enlarge the number of broken feature sequences used in training. 
As the average number of instances per day was found to be 40, we generated synthetic sequences of length 40 by merging normal and defective features to mime breaking within one day. We selected one sequence of each class randomly and filled the new series by fading. 
During most breakage scenarios the feature values rose according to a sigmoidal shape.
Therefore, fading is performed using a sigmoid function, which was randomly scaled and shifted along the y-axis. Equation~\ref{equ_timeSeries} describes how the synthetic values $x$ for the time stamps $j~\in~[0,40[$ are calculated based on values of randomly chosen normal ($normal[j]$) and broken ($broken[j]$) time series at element $j$. Equation~\ref{equ_sigmoid} presents the used sigmoid function $p[j]$.
\begin{equation}
x[j] = (1 - p[j]) \cdot normal[j] + p[j] \cdot broken[j]
\label{equ_timeSeries}
\end{equation}
\begin{equation}
p[j] = \frac{1}{1 + \exp(- \frac{j-\mu}{\sigma})}
\label{equ_sigmoid}
\end{equation}
The value for scaling factor $\sigma$ is randomly chosen out of range $[0.2,1]$, whereas $\mu$ carries the translation factor which can have values between $-13.3$ and $13.3$. The allowed ranges were determined experimentally.
This is performed for all four features (CNL, CSP, SSR, CSI) individually using the same sigmoid function.

\subsection{Classification}
We applied the following four different time series classification methods to the preprocessed data and compared them. Thus, we determined the best suiting of the four models in order to predict which coils will change their state from normal to defective. All parameters were determined using hyper-parameter tuning. Thus, the model with the lowest validation loss was chosen after random initialization of weights and optimization following Adam's proposal~\cite{kingma2014adam}.

A leave-several-coils-out cross validation was performed, where the respective non-testing coils were stratified split into 70\% training and 30\% validation data. Thus, we can assure that only sequences from distinct coils were used in model training and testing. 

\subsubsection{Fully Convolutional Neural Networks}
Following the approach from Wang et al.~\cite{wang2017time}, we implemented a Fully Convolutional Neural Network (FCN) which is composed of three convolutional blocks. All blocks perform three operations, each. The first block contains a convolution with 128 filters of length eight followed by a
batch normalization~\cite{ioffe2015batch} using 256 filters with a filter length of five in order to speed up convergence. Its result is sent to a ReLu activation function consisting of 128 filters where each has a length equal to three. After the calculation of the third and last convolution block, the average over the complete time is calculated. This step is comparable to a Global Average Pooling (GAP) layer. Finally, the GAP layer's output is fully connected to a traditional softmax classifier. In order to maintain the exact length of the time series throughout the performed convolutions, zero padding and a stride equal to one were used in every convolution step. The FCN does not contain any pooling to prevent overfitting nor a regularization operation.

\subsubsection{Residual Network}
Moreover, we set up a ResNet proposed by~\cite{wang2017time} built out of three residual blocks. Each block is composed of three convolutions. The convolution result of each block is added to the shortcut residual connection (input of each residual block) and is then fed to the following residual block. For all convolutions the number of filters used is set to 64, with the ReLU activation function followed by a batch normalization operation. Within the first block the filter length is set to eight, in the second one to five and in the third to three. The three residual blocks are followed by a GAP layer and a softmax classifier whose number of neurons is equivalent to the number of classes in the data set. The main characteristic of ResNets and the difference to usual convolutions is the linear shortcut between input and residual blocks which allows the flow of the gradient directly through these linear connections. Therefore, training is simplified as the vanishing gradient effect is reduced.

\subsubsection{Time Convolutional Neural Networks}
As an alternative, we implemented a Time Convolutional Neural Network~\cite{wang2017time}(TCNN) constructed of two consecutive convolutional layers with six and twelve filters. A
local average pooling operation of length three follows the convolutional layers. Sigmoid is used as the activation function. The network's output results in a fully connected layer, where the number of neurons is in our case two as we focus on two classes in our data set, normal and broken coil.

\subsubsection{Long Short-Term Memory}
The forth method we applied is a LSTM network. It contains two convolutional layers without padding operations. Therefore, the sequence length decreases with every convolution. Each convolutional layer is preceded by local average pooling operation of length three and a dropout operation with a rate equal to 0.2 to prevent overfitting~\cite{srivastava2014dropout}. After the construction of convolutional layers, average pooling and dropout operations, two LSTM layers with 32 units and a $\tanh$ activation function follow. A dense layer and a sigmoid classifier whose number of neurons is equivalent to the number of classes in the data set complete the model structure.

\section{Results}
We applied our algorithms on the given time series data consisting of 97.8\% normal and 2.2\% broken samples first without adding synthetic sequences. Table~\ref{tab:TSC_performance} holds the performance measures accuracy, precision, recall and F-score for the different models where we underlined the best scores. We see that the four models perform very similar in terms of accuracy only ranging from 97.43\% for ResNet up to 98.33\% reached by LSTM. Furthermore, LSTM classifies coil sequences most precisely with 90.10\% and delivers 84.16\% recall. This results also in the best F-score for LSTM reporting 86.43\%. Next to the performance measures, the confusion matrices are given holding True Negative (TN), False Positive (FP), False Negative (FN), and True Positive (TP) rates.

\begin{table}[th]
	\caption{Average prediction performance measures and confusion matrix of the applied TSC methods after 10-fold cross-validation.}
	\centering
	\begin{tabular}{lcccc|cccc}
		\hline
		{\%}  & {Accuracy}  & {Precision} & {Recall} & {F-Score} & {TN} & {FP} & {FN} & {TP} \\ 
		{FCN} & 97.53 &	75.54 &	\underline{85.68} &	75.66 & 97.92 &	2.08 &	14.31 &	85.69\\
		{ResNet} & 97.43 &	75.73 &	83.75 &	73.98 & 97.72 &	2.28&	16.23 &	83.77 \\
		{TCNN} & 97.72 &	82.56 &	68.23 &	74.16 & 99.33 &	0.67 &	29.35 &	70.65\\
		{LSTM} & \underline{98.33} &	\underline{90.10} &	84.16 &	\underline{86.43} & 98.67 &	1.33 &	15.83 &	84.17\\
		\hline
	\end{tabular}
	
	\label{tab:TSC_performance}
\end{table}
Due to the highly imbalanced class distribution, we generated synthetic data to increase the number of broken samples in the training set. As LSTM achieved the best performances in all four measures, we applied LSTM on three different augmentation degrees. Table~\ref{tab:TSC_augmentation_LSTM} presents the performance measures and confusion matrices. For comparison, results for the original data set and the augmented datasets containing 2.4\% and 2.6\% broken instances are listed.

\begin{table}[th]
	\caption{Average number of broken instances, prediction performance measures and confusion matrix after 10-fold cross-validation. Different results were achieved with LSTM for different degrees of synthetically increased numbers of broken sequences during model fitting.}
	\centering
	\begin{tabular}{c|cccc|cccc}
		\hline
		{Broken instances}  & {Accuracy }  & {Precision} & {Recall} & {F-Score} & {TN} & {FP} & {FN} & {TP} \\ 
		{576 (2.2\%)} & 98.33 &	\underline{90.10} &	\underline{84.16} &	\underline{86.43} & 98.67 &	1.33 &	15.83 &	84.17\\
		{641 (2.4\%)} & 97.67 & 83.32 & 71.49 & 76.28& 99.18 & 0.82 & 28.51 & 71.49\\
		{706 (2.6\%)} & \underline{98.38} & 88.02 & 82.29 & 83.59& 98.75 & 1.25 & 17.71 & 82.29\\
		\hline
	\end{tabular}
	\label{tab:TSC_augmentation_LSTM}
\end{table}

\section{Discussion}
For the task of predicting normal and broken Head/Neck coils based on collected image features over time, we found the presented LSTM resulting in highest accuracy of 98.33\% and F-score of 86.43\%. We showed that LSTM, FCN and ResNet misclassified only few defective coils as normally functioning and thus, received similar recall values. Although ResNet has a deep and flexible architecture, LSTM enabled to classify least sequences within the individual normal coils incorrectly.
We explain those results with the mixture of cases in our training data. We observed sequences with very current irregularities as well as features that were collected longer ago. Thus, LSTM considers both cases and combines long and short term instances beneficially.
Augmentation of time series data for training by combining normal and defective sequences did not increase the average prediction performance measures significantly. This shows the model did not gain any additional information from the synthetic data which would ease the classification task of the test data. We did not find research applying time series classification methods to image features in order to detect hardware failures. Thus, we consider us to be the first using sequential data for hardware failure prediction while achieving high performance.

We only picked four models for comparison. In a next step, more models should be considered and tested on a larger data set.
Moreover, in future work also other augmentation possibilities should be considered. As coils brake because of different reasons reporting different variations in  image features, the applied sigmoid function might not reflect real world scenarios, exhaustively.
Furthermore, time series classification should be also applied to other MRI hardware in order to determine the generality of our algorithm trained and tested for Head/Neck coils.

\bibliographystyle{bvm2020}

\bibliography{2772}
\marginpar{\color{white}E\articlenumber} 
\end{document}